\documentclass[11pt,letterpaper]{article}

\usepackage[utf8]{inputenc} % allow utf-8 input
\usepackage[T1]{fontenc}    % use 8-bit T1 fonts
\usepackage{lmodern}
\usepackage{hyperref}       % hyperlinks
\usepackage{url}            % simple URL typesetting
\usepackage{booktabs}       % professional-quality tables
\usepackage{amsfonts}       % blackboard math symbols
\usepackage{nicefrac}       % compact symbols for 1/2, etc.
\usepackage{microtype}      % microtypography
\usepackage{xcolor}         % colors
\usepackage{geometry}
\usepackage{graphicx}
\usepackage{amsthm,amsmath}

\usepackage{dingbat}
\usepackage{cleveref}
\newtheorem{theorem}{Theorem}
\newtheorem{lemma}[theorem]{Lemma}

\newtheorem{corollary}[theorem]{Corollary}
\newtheorem{definition}[theorem]{Definition}
\newtheorem{proposition}[theorem]{Proposition}

\usepackage{authblk}

\def\calP{\mathcal{P}}
\def\P{\mathbb{P}}
\def\E{\mathbb{E}}

\def\like{\text{\leftthumbsup}}
\def\dislike{\text{\rightthumbsdown}}

\title{Dynamic pricing with Bayesian updates from online reviews
\thanks{An extended abstract of a  preliminary version of this paper has been presented in the NeurIPS Workshop on ML for Economic Policy in 2020.}
}
\date{April 2024}
 \author[1]{José Correa}
 \author[2]{Mathieu Mari}
 \author[3]{Andrew Xia}
 \affil[1]{Universidad de Chile, Chile. {\tt correa@uchile.cl}.}
 \affil[2]{LIRMM, Université de Montpellier, CNRS, Montpellier, France. {\tt mathieu.mari@lirmm.fr}.}
 \affil[3]{Flagship, USA. {\tt axia@alum.mit.edu}.}

\begin{document}

\maketitle

\begin{abstract}
When launching new products, firms face uncertainty about market reception. Online reviews provide valuable information not only to consumers but also to firms, allowing firms to adjust the product characteristics, including its selling price. In this paper, we consider a pricing model with online reviews in which the quality of the product is uncertain, and both the seller and the buyers Bayesianly update their beliefs to make purchasing \& pricing decisions. We model the seller's pricing problem as a basic bandits' problem and show a close connection with the celebrated Catalan numbers, allowing us to efficiently compute the overall future discounted reward of the seller. With this tool, we analyze and compare the optimal static and dynamic pricing strategies in terms of the probability of effectively learning the quality of the product. 

% \noindent\emph{Problem definition:} 
% In this paper, we consider a pricing model with online reviews in which the quality of the product is uncertain, and both the seller and the buyers Bayesianly update their beliefs to make purchasing \& pricing decisions. The goal is to find and compare the optimal dynamic and static pricing policies.

% \vspace{2ex}\noindent\emph{Methodology/results:} 
% We model the seller's pricing problem as a basic bandits' problem and show a close connection with the celebrated Catalan numbers, allowing us to efficiently compute the seller's overall future discounted reward. With this tool, we analyze and compare the optimal static and dynamic pricing strategies in terms of the probability of effectively learning the quality of the product. 

% \vspace{2ex}\noindent\emph{Managerial implications:} 
% When launching new products, firms face uncertainty about market reception. Online reviews provide valuable information not only to consumers but also to firms, allowing firms to adjust the product characteristics, including its selling price. Though stylized, our model and results can help firms understand the connections between the learning process and dynamic pricing policies. 
% \vspace{5ex}
\end{abstract}

\section{Introduction}

As a key part of modern online platforms, online decision-making plays a crucial role in a variety of settings, particularly related to the Internet. Two landmark examples that have been widely studied are \emph{dynamic pricing} and \emph{online reviews}. Online review systems constitute powerful platforms for users to get informed about the product and for the firm to understand how a given market is receiving the product. The study of these systems has been vast for the last two decades \cite{DLP03,HL04}, and more recently, modeling simple like/dislike reviews as bandits problems have become standard \cite{AMMO19,BS18,BP13,MA19, SJB19, VMCS18}. Dynamic pricing, on the other hand, is an active area of research in economics, computer science, and operations research \cite{MSA19, PS17}, and has become a common practice in several industries such as transportation and retail. 

There has been a growing interest in combining the two areas as a way to design more effective pricing mechanisms that gather information from current reviews to update prices and make the product more attractive \cite{CIMS17,IMSZ19,SVZ19}. In particular, \cite{CIMS17} considers social learning with non-Bayesian agents in a market with like \& dislike reviews, and the resulting pricing decision of a monopolist. \cite{SVZ19} considers a setting when the volume of sales is large and optimizes revenue via fluid dynamics and ODEs. The general idea here is to use online reviews to update the market's belief about the quality of the product, thus influencing its pricing. However, the complexity of many models limits their practicality as the Bayesian updating of beliefs becomes intractable. 

In this paper, we continue on this line of research. We consider a straightforward model that precisely determines the optimal pricing strategies from information elicited by online reviews. The main message of this paper is to show that online reviews not only influence how successful a product is but also help find more effective dynamic pricing strategies. These dynamic pricing strategies lead to more efficient allocations. Dynamic pricing with online reviews gives a foundation for the common practice of temporarily pricing a product below its production cost, leading to short-term revenue losses. These losses come with a potential boost in future purchases, even at a higher price, and ultimately may lead to more revenue. 

\section{Preliminaries}

\subsection{Our model.} We first give a general description of our model and then define its features precisely. We defer a discussion of the most closely related models \cite{CDKS16,CIMS17,IMSZ19,SVZ19}
 to \Cref{related}.

Consider a seller marketing a new product. Neither the seller nor the buyers are informed about the quality of the product but only receive a public signal, representing the prior probability of the quality of the product. Based on this estimation and the product price, buyers, that arrive in an online fashion, estimate their expected utility and decide to buy the product or not. If no purchase occurs, the buying process terminates. Otherwise, after buying, the buyer experiences the product and may like it or not; in both cases, he submits an online review with either a like or dislike. These reviews allow following users to update their priors on the quality of the product. 

\paragraph{\bf Product.} As it is common in the literature \cite{AMMO19, IMSZ19} we assume that the product may be either \emph{good} or \emph{bad}. A \emph{good} product will be liked by a user with probability $p$ (so that roughly a fraction $p$ of the market is satisfied with the product), while a \emph{bad} product will be liked by a user with probability $q<p$. We also assume that the product has a fixed production cost $c$ per unit. To avoid trivial cases, we assume in the paper that $q<c<p$. Finally, the product has a price $\pi$ that may evolve along the process. The \emph{prior} probability on the quality of the product, denoted by $x$, is the probability that the product is good. All $p,q$, and $x$ are common knowledge.
\paragraph{\bf Buyers.} The market is composed of an infinite stream of users arriving at times $t=0,1,2,\ldots$, which are offered the product at a certain, possibly time-dependent, price $\pi$. Upon receiving the offer, a user (which we assume risk-neutral) evaluates his expected utility for buying the product.  Initially, since the first buyer's prior is $x$, his expected utility can be evaluated as $xp+(1-x)q$. If this quantity exceeds the current price $\pi$, then he decides to buy the product. 
\paragraph{\bf Priors.} After experiencing the product, the buyer submits an online review in the like/dislike format. If the buyer likes ($\like$) the product then, given a current prior $x$, we update the prior to $L(x)$ as follows:
\begin{align*}
    L(x)&:=\mathbb{P}_x(\text{good }| \like ) = \frac{\mathbb{P}_x(\like | \text{good})\mathbb{P}_x(\text{good})}{\mathbb{P}_x(\like)}
    = \frac{x \cdot p}{x\cdot p + (1-x)\cdot q}.
\end{align*}
Similarly, given a dislike ($\dislike$), we update the prior as follows: 
\begin{align*}
    D(x):&=\mathbb{P}_x(\text{good } | \dislike) = \frac{\mathbb{P}_x( \dislike | \text{good})\mathbb{P}_x(\text{good})}{\mathbb{P}_x(\dislike)}
    = \frac{x\cdot (1-p)}{x\cdot (1-p) + (1-x)\cdot (1-q)}.
\end{align*}
  
In particular, the prior increases after a like, and decreases after a dislike: $D(x)<x<L(x)$.  An interesting feature of our model is that the updated prior after a sequence of likes and dislikes only depends on the number of reviews and not the sequence of reviews. This is in contrast with most models of online reviews, and it allows both the seller and the users to update their beliefs based solely on these figures. 

\begin{lemma}
    Given a prior $x$, the updated prior after a sequence of $\ell$ likes and $d$ dislikes does not depend on the order and this value is $x_{\ell,d}=\frac{xp^\ell(1-p)^d}{ xp^\ell(1-p)^d + (1-x)q^\ell(1-q)^d}$. Furthermore, the probability of each such sequence only depends on $\ell$ and $d$.
    \label{lemma:xwl}
\end{lemma}
\begin{proof}
Let us start by calculating $D(L(x))$, the value of the updated prior after seeing a like and then a dislike:
\begin{align*}
    D(L(x)) &= \frac{(1-p)L(x)}{(1-p)L(x) + (1-q)(1-L(x))} 
    =  \frac{(1-p)\frac{p \cdot x}{p\cdot x + q\cdot (1-x)}}{(1-p)\frac{p \cdot x}{p\cdot x + q\cdot (1-x)} + (1-q)(1-\frac{p \cdot x}{p\cdot x + q\cdot (1-x)})}\\
    &= \frac{p(1-p)\cdot x}{p(1-p)\cdot x + q(1-q)\cdot(1-x)}
\end{align*}

Clearly, this expression when swapping $p$ (resp. $q$) with $1-p$ (resp. $1-q$) is unchanged, thus $L(D(x)) = D(L(x))$. 

The proof of the formula for $x_{\ell,d}$, with $\ell$ likes and $d$ dislikes, uses induction and works similarly.

We now show that the probability of observing a given sequence of likes and dislikes is independent of the order. To see this, we calculate 
\begin{align*}
\mathbb{P}_{L(x)}(\dislike)&\cdot \mathbb{P}_x(\like) 
 = \left( L(x)(1-p) + (1-L(x))(1-q) \right) \cdot \left( xp + (1-x)q \right)\\
& =\left( \frac{xp}{x p + (1-x) q}(1-p) + (1-\frac{x p}{x p + (1-x) q})(1-q) \right)
\cdot \left( xp + (1-x)q \right)\\
&      = xp(1-p) + (1-x)q(1-q).
      \end{align*}
And we easily get the same expression for $\mathbb{P}_{D(x)}(\like)\cdot \mathbb{P}_x(\dislike)$. The result follows by induction.  
\end{proof}

{\bf Seller's problem. }
A basic problem faced by the seller is to find an optimal pricing strategy. One alternative for the seller is to adopt a \emph{static} price: the price $\pi$ offered to each consumer is fixed at the beginning and can not be changed. In this situation, users will buy the product as long as the current prior $x$ satisfies $xp+(1-x)q\ge \pi$ and yields positive expected value. Whenever $xp+(1-x)q < \pi$, the process will stop forever. In other words, the sales process will continue until the prior reaches $x_{\min}=\frac{\pi-q}{p-q}$. The \emph{local reward} for the seller for each sale is $R(x)=\pi-c$. 
Given the prior $x$, the seller, who discounts the future at rate $\delta$, can express her \emph{expected revenue} or \emph{expected global reward} recursively as $V(x)=0$ if $x<x_{\min}$, and otherwise
\begin{align}
    V(x)=R(x) &+ \delta\cdot\mathbb{P}_x(\text{\like})\cdot V(L(x))
    + \delta\cdot \mathbb{P}_x(\dislike)\cdot V(D(x)).
    \label{eq:future_revenue}
\end{align}
Finally, the seller will optimize this function over the values of $\pi\in (c,xp+(1-x)q]$.\footnote{Clearly $\pi$ has to be at least $c$ for the revenue to be positive. Also, if $\pi>xp+(1-x)q$, then no user will ever buy, and then the revenue is zero.}

On the other hand, the seller may opt for a \emph{dynamic} pricing approach. In this setting the price may be adjusted according to the current reviews the product has received. To maximize revenue, the seller will just make the user indifferent to purchase whenever she decides to continue selling the product, offering the product at time $t$ with prior $x$ at exactly $\pi_t= xp+(1-x)q$. Thus, with dynamic prices, the local reward for the seller after each sale depends on the prior with value $R(x)=xp+(1-x)q-c$.\footnote{Note that, slightly abusing notation, we always use $R(x)$ for the local reward, although its value depends on whether we are considering static or dynamic pricing.} 
The decision of when to stop, however, becomes more involved. Although at times $xp+(1-x)q<c$ may hold, and therefore in the next sale the seller will incur in a loss, it may still be worth to continue selling. The reason for this relies on the information gain provided by one more sale and the impact this information has on future purchases. In this scenario, given a prior $x$, to express her expected revenue, the seller can either decide to stop selling, and her reward is thus $V(x)=0$; or can decide to continue selling, and then her revenue can be expressed as in equation \eqref{eq:future_revenue}. To maximize her profit, she will pick the best of these two options, and thus her \emph{expected global reward} satisfies the following recursive equation: 
\begin{align}
   V(x) = \max \Big(0, R(x) &+ \delta\cdot\mathbb{P}_x(\text{\like})\cdot V(L(x))
    + \delta\cdot \mathbb{P}_x(\dislike)\cdot V(D(x))\Big), \label{eq:value}
\end{align}
where $R(x)=xp+(1-x)q-c$. 

Therefore, with dynamic prices, the seller stops selling when the current prior $x$ satisfies $x<x^*$, 
where $x^*$ is defined as the largest that $V(x^*)=0$, (where $V$ is the solution of \eqref{eq:value}). This value determines the stopping time of the seller under dynamic pricing.

Thus, the expected global reward in both pricing scenarios can be expressed with the same recursion:
\begin{equation}
V(x)=\begin{cases}
0,\hspace{265pt}\,\, \text{ if }x<x_\text{stop}\\
R(x) + \delta\cdot\mathbb{P}_x(\text{\like})\cdot V(L(x))+ \delta\cdot \mathbb{P}_x(\dislike)\cdot V(D(x)), \hspace{30pt}\text{ otherwise.}
\end{cases}
\label{eq:recursive_process}
\end{equation}
Here, for static pricing we have $x_\text{stop}=x_{\min}=\frac{\pi-q}{p-q}$ and $R(x)=\pi-c$, while for dynamic pricing we have $x_\text{stop}=x^*$ and $R(x)=xp+(1-x)q-c$. 

\subsection{Our Results} In this paper we study dynamic and static pricing with online reviews in detail. We first show that we can formulate the underlying dynamic process as a simple multi-armed bandit problem. Perhaps surprisingly, this problem appears to be unexplored in the literature. The core of the paper is devoted to the computation of the stopping prior $x^*$ in the dynamic price setting and the computation of the global expected reward in both settings. For this, we propose two approaches. 

We first present a fast dynamic programming approach in  \Cref{sec:dyn_prog} that computes an approximation of $x^*$ and an approximation of the value of the global expected reward in both pricing scenarios. The backside of this fast heuristic is that we are unable to provide a good guarantee on the quality of the solutions produced. 

To resolve this issue, in  \Cref{section:combinatorial_approach}, we take a combinatorial approach using the classic Gittins index \cite{GJ79}, and we explicitly determine the optimal underlying stopping time and the global excepted rewards. Specifically, we obtain a closed-form formula for these quantities. To this end, we uncover a connection with the classic Catalan numbers. With this, we are able to compute an arbitrarily good approximation for $x^*$ and the global expected reward in both pricing scenarios. 

In \Cref{sec:comparison}, we study, for each pricing strategy, the probability of achieving full efficiency. Note that this fully efficient situation occurs when the product is good and is sold forever\footnote{This is the best possible situation for the whole market, considering both the seller and the buyers.}, i.e., the product is good, and the market learns this fact. To this end, we exploit that the stochastic process governing the prior updates is a martingale, and we can thus use the optional stopping theorem. 

We finally discuss in \Cref{sec:extension} an extension of the model in which the quality of the product is not restricted to be {\em good} or {\em bad}, but it can take arbitrary values in a set $Q\subseteq [0,1]$. For this more general model we note that the history independence property still holds and that, in essence, our dynamic programming approach can still be used to compute the seller's revenue. We additionally observe that when the set of possible product qualities $Q$ is a continuous set and the initial prior quality distribution is sufficiently smooth, the dynamic program is very effective since already a short sequence of reviews gives a very good estimate about the true quality of the product. 

\subsection{The Bandits Connection} We can model our dynamic pricing problem using a bandit framework. To the best of our knowledge, an optimal strategy for that bandit problem was previously unknown. 
Imagine there is a single slot machine (bandit), which could either be a \emph{good} or \emph{bad} machine. 
This machine costs $c$ to play and will yield a return of $1$  
with some probability and $0$ 
with some other probability.\footnote{Again, to avoid trivial cases we assume that $0 \leq q < c < p \leq 1$.} 
If the machine is good, it has a fixed, known probability of $p$ of returning $1$, while if it is bad it has known probability of $q$ of returning $1$. We have a prior $x$ that the machine is good, however we do not know for sure if the machine is good or not. Thus, given a prior $x$, the expected earning of a single pull is $xp + (1-x)q - c$. Finally, we discount the future at rate $\delta$, so that the value of earnings in time $t$ is discounted by a factor of $\delta^t$. As we play, we update our prior using Bayes rule and the problem we consider is that of determining the prior, $x^*$, under which we should stop playing.

The correspondence between the dynamic pricing problem and the bandits problem is straightforward. 
The only apparent difference in the problems comes from the available information. In bandit problems, we typically assume that we have the whole history of pulls, whereas in the dynamic pricing problem, we only want to assume that the users get to see the number of likes and dislikes the product has received so far. However, as observed in  \Cref{lemma:xwl} this is not an issue since the updated prior after a number of reviews only depends on the number of likes and dislikes and not on the sequence itself.

\section{Related models}\label{related}
Our model adds to the literature on pricing with online reviews. Most of these existing models try to make simplifying assumptions, but even with these, they end up being extremely difficult to solve and analyze. On the contrary, the model we consider is relatively simple (though still realistic) and can be solved exactly. In what follows we present a more detailed comparison between our model and the closest related work in the literature. 

First, let us discuss the model of Crapis et al. \cite{CIMS17}. As in our model, they consider an infinite stream of buyers purchasing a product. The quality $p$ of the product is also initially unknown and can take values in an interval (so, in this sense, it is similar to our extended model). As opposed to our model (and actually to most models in social learning) they consider heterogeneous buyers. The utility of buyer $i$ is given by $u_i = \alpha_i q - p,$ where p is the price charged by the seller, and the $\alpha_i$ are i.i.d. random variables
drawn from a known distribution $F$. They also consider that buyers arrive according to an independent Poisson process, but this does not significantly affect the results. Initially, all buyers have some common prior on the quality of the product, $q_0$. Consumers report likes and dislikes depending on whether their utility was nonnegative, taking into account the true quality of the product (which was discovered upon buying it). The information available to the buyer upon making his purchasing decision includes reviews made by all of his predecessors and knowledge about the order in which they acted. The seller's problem is to find the price maximizing her discounted expected revenue. We remark that the information structure here is quite involved, so, in order to tackle the seller's problem, the authors make some simplifying assumptions and resort to mean-field approximations.

In subsequent work, Ifrach et al. \cite{IMSZ19} further refine the latter model. In particular, they reduce the possible \emph{qualities} that the product can take to two possible values: \emph{high} and \emph{low}. Accordingly they modify the form of the utility of buyers to take an additive form. Namely, a buyer's utility equals to the quality of the product, minus the price paid, plus the buyer's type which are represented by i.i.d. random variables. Again, a like/dislike represents whether the buyer's utility was positive/negative. With these assumptions, the information structure gets somewhat simplified, although it is still quite complex. In this paper, as in ours, the authors additionally assume that the product has a cost $c$ and observe that this cost plays an important role in the optimal dynamic pricing policy, and on whether consumers ultimately learn the true quality of the product. Closely related to the model of Ifrach et al., is that of Acemoglu et al., \cite{AMMO19}. Similar to our work they consider static and dynamic pricing strategies for the seller and give conditions under which asymptotic learning occurs, and at which speed. A key distinction in Acemoglu et al.'s model refers to whether buyers have access to summary statistics or to the full history. We note that in our model setup (which is different) this distinction is unnecessary as we prove that both settings coincide.

Very recently, Shin et al. \cite{SVZ19} take a slightly different approach to the problem. They consider a finite horizon model in which at each point in time a new buyer shows up. In their model buyers are homogeneous and the utility of buyer $i$ is given by $u_i=q_i-p_i$, where $q_i$ is the experienced quality of the product and $p_i$ the price paid. This paper takes a different approach to modeling the quality of the product. This is assumed to be a Gaussian random variable of mean $\mu$ and standard deviation 1. However, only the seller known the mean while the buyers' use online reviews to learn this parameter. Another different feature of this paper is that online reviews may take numerical values beyond like/dislike (say star rating). As the authors note, the resulting dynamic pricing problem is extremely hard to solve analytically so they end up looking at asymptotically optimal pricing policies.

Another related paper is the work of Chawla et al. \cite{CDKS16}. This paper, however, follows a different language that makes the comparison slightly more difficult. The authors consider a buyer who repeatedly interacts with a seller. The buyer does not know his valuation, and every time he purchases the product, he updates his valuation $V_t$. This value can be thought of as the prior on the quality in our model. The evolution of $V_t$ is assumed to follow a martingale, which naturally holds in our Bayesian updating. However, Chawla et al. make some assumptions about the variance and the step size of the random process (that then affect the main results). Another difference is that this paper does not consider the cost of the product, which plays an important role in our and other dynamic pricing models. The main result of Chawla et al. is that, under suitable conditions, a simple pricing strategy in which the product is given away for free up until some point, and then a fixed price is used, recovers a good fraction of the optimal revenue.

\section{A Dynamic Programming Approach}\label{sec:dyn_prog}

In this section, we propose a fast dynamic programming approach to compute an approximation of the threshold prior $x^*$ and the global expected reward in the dynamic pricing scenario. Then, we explain how to adapt this algorithm to compute an approximation of the global expected reward in the static price scenario. 

Let us denote the set of possible updated priors by $\calP(x):=\{x_{\ell,d} \mid \ell,d \ge 0\}$. We say that a set $S\subseteq [0,1]$ is \emph{discrete} if all elements of $S$ have a neighborhood that contains no other elements of $S$. The next result characterizes when the set of possible updated priors is discrete. We note that whenever it is not, then $\calP(x)$ is dense in $[0,1]$.

\begin{proposition}
The space of possible updated priors $\calP(x)$ is discrete if and only if $$\frac{\log(\frac{p}{q})}{\log(\frac{1-q}{1-p})}\in \mathbb{Q}.$$ 
\label{claim:rational}
Furthermore, whenever $\calP(x)$ is not a discrete set, it is dense in $[0,1]$.
\end{proposition}

\begin{proof}
   First, assume that 
    $$\gamma:=\frac{\log(\frac{p}{q})}{\log(\frac{1-q}{1-p})}=\frac{a}{b},$$ where $a$ and $b$ are two integers such that $gcd(a,b)=1$. This is equivalent to $\left(\frac{1-q}{1-p}\right)^a=\left(\frac{p}{q}\right)^b$.  We show that $\calP(x)$ is discrete, by arranging its elements as an infinite increasing sequence $(x_i)_{i\in \mathbb{Z}}$ such that $x_0=x$ and for all $i\in \mathbb{Z}$ we have $L(x_i)=x_{i+a}$ and $D(x_i)=x_{i-b}$. 
    
   We first show that the updated prior after a sequence of $b$ likes and $a$ dislikes is unchanged, i.e. $x_{b,a}=x_{0,0}$. 
    Indeed, by \Cref{lemma:xwl}, we have $$x_{b,a}=\frac{xp^b(1-p)^a}{ xp^b(1-p)^a + (1-x)q^b(1-q)^a}=\frac{x}{x+ (1-x)(\frac{q}{p})^b(\frac{1-q}{1-p})^a}=x.$$ 
    
    Then, for any $i\in \mathbb{Z}$, we can set $x_i:=x_{\ell,d}$ where $\ell,d$ is any pair of integers such that $\ell\cdot a-d\cdot b=i$. The sequence $(x_i)_{i\in \mathbb{Z}}$ is strictly increasing because $a\ell-bd>a\ell'-bd'$ if and only if $(\frac{q}{p})^\ell(\frac{1-q}{1-p})^d>(\frac{q}{p})^{\ell'}(\frac{1-q}{1-p})^{d'}$, i.e., if and only if $x_{\ell,d}>x_{\ell',d'}$. 

    Now, we show that $\mathcal{P}(x)$ is dense in $[0,1]$ when $\gamma$ is irrational. Since $x_{\ell,d}$ can be re-written as 
$$
x_{\ell,d}=\dfrac{x}{x+(1-x)\exp(\log(p/q)\cdot(d\gamma^{-1}-\ell))}
$$
and  $y\mapsto \dfrac{x}{x+(1-x)\exp(\log(p/q)\cdot y)}$ is a continuous one-to-one function from $\mathbb{R}$ to $(0,1)$ of bounded derivative, it is enough to show that the set $\{d\gamma^{-1}-\ell\mid \ell,d\ge 0\}$ is dense in $\mathbb{R}$. To show that, it is in fact sufficient to show that the set $H=\{h(d), d\ge 0\}$ is dense in $[0,1]$, where $h(d):=d\gamma^{-1} -\lfloor d\gamma^{-1}\rfloor$. 

The first step to prove that $H$ is dense in $[0,1]$ is to observe that it is infinite. Otherwise, there would exist two distinct integers $d$ and $d'$ such that $h(d) = h(d')$, which would contradict our assumption that $\gamma$ is irrational. 

Next, let $c\in [0,1]$ and $\epsilon>0$. We show that that there exists $h\in H$ such that $|h-c|<\epsilon$. Since $H$ is infinite and compact, there exist two elements of $H$ that are at a distance less than $\epsilon$ from each other, i.e., there are two integers $d$ and $d'$ with $d>d'$ such that $0<|h(d)-h(d')|<\epsilon$. First, suppose that $h(d)>h(d')$. Then, $h(d-d')=h(d)-h(d')<\epsilon$ and the element
$$
h\left(\left\lfloor\frac{c}{h(d-d')} \right\rfloor(d-d')\right)=\left\lfloor\frac{c}{h(d-d')} \right\rfloor h(d-d')
$$
is in $H$ and is at distance less than $\epsilon$ from $c$, what we wanted to find. In the other case we have $h(d)<h(d')$ and then $1-\epsilon<h(d-d')<1$. In that case, the element 
$$
h\left(\left\lfloor\frac{1-c}{1-h(d-d')}\right\rfloor(d-d')\right)=\left\lfloor\frac{1-c}{1-h(d-d')}\right\rfloor h(d-d')
$$ is in $H$ and is at distance $\epsilon$ from $c$. 
This completes the proof. 
\end{proof}

\begin{corollary} Let $a$ and $b$ be two integers such that $gcd(a,b)=1$ and assume that 
$$\frac{\log(\frac{1-q}{1-p})}{\log(\frac{p}{q})}=\frac{a}{b}.$$ 
Then there exists an infinite increasing sequence $(x_i)_{i\in \mathbb{Z}}$ such that for all $i\in \mathbb{Z}$ we have $L(x_i)=x_{i+a}$ and $D(x_i)=x_{i-b}$.
\end{corollary}

In this setting, we can re-write \cref{eq:recursive_process} as 
$V(x_i)=0$ if $x<x_\text{stop}$ and otherwise: 
$$V(x_i)=R(x_i) + \delta\mathbb{P}_{x_i}(\text{\like})V(x_{i+a})
    + \delta \mathbb{P}_{x_i}(\dislike)V(x_{i-b})$$
{\bf The algorithm.} 
To apply a dynamic programming approach, let us fix an $\epsilon>0$ that corresponds to the threshold between the precision of the solution returned and the running time: the smaller $\epsilon$, the greater the precision and the running time. We note that this approach is not mathematically rigorous since its validity requires smoothness conditions on the function $V(\cdot)$ that do not follow from 
\cref{eq:recursive_process}. In the next section we  derive a formal approach while now we continue with this approach that is computationally tractable.

First, assuming that $V$ is differentiable at $x=1$ we get from \cref{eq:recursive_process} that $V(1)=\frac{p-c}{1-\delta}$ and $V'(1)=\frac{p-q}{1-\delta}$, as the following proposition shows.

\begin{proposition}
Assume that the solution $V$ of \cref{eq:value} is differentiable at $x=1$, then we have $V(1)=\frac{p-c}{1-\delta}$ and $V'(1)=\frac{p-q}{1-\delta}$.
\label{claim:V'(1)}
\end{proposition}
\begin{proof}
    If $x=1$, then the product must be good; therefore the expected global reward is $$V(1)=\sum_{t\ge 0} (p-c)\delta^t=\frac{p-c}{1-\delta}.$$ Now fix a small $\epsilon>0$. Assume that the function $V(x)$ admits a derivative in $x=1$, we can write $V(1-\epsilon)=V(1)-\epsilon V'(1) + o(\epsilon)$. With \cref{eq:value} we have for $x=1-\epsilon$ close to $1$:
    \small
    \begin{align*}
        V(1-\epsilon)&=R(1-\epsilon) + \delta\cdot\mathbb{P}_{x}(\like)\cdot V(L(1-\epsilon))
         \delta\cdot \mathbb{P}_{x}(\dislike)\cdot V(D(1-\epsilon))\\
        &= R(1-\epsilon) + \delta(p - \epsilon(p-q))(V(1-\frac{q}{p}\epsilon + o(\epsilon))) \\ 
        &\hspace{20pt}+ \delta((1-p) + \epsilon(p-q))(V(1-\frac{1-q}{1-p}\epsilon + o(\epsilon)))\\
        & =R(1-\epsilon) + \delta(p - \epsilon(p-q))(V(1)+\frac{q}{p}\epsilon V'(1)+ o(\epsilon)) \\
        &\hspace{40pt}+ \delta((1-p) + \epsilon(p-q))(V(1)+\frac{1-q}{1-p}\epsilon V'(1)+ o(\epsilon))\\
        & =R(1-\epsilon) + \delta V(1) - \epsilon\delta V'(1)+o(\epsilon)\\
        & =p-c + \delta V(1) + \epsilon\left( -(p-q) + \delta V'(1)\right)+o(\epsilon)\\
        &=V(1)-\epsilon V'(1) + o(\epsilon).
    \end{align*}
    \normalsize
    Thus, $V'(1)=\frac{p-q}{1-\delta}$.
\end{proof}

With \Cref{claim:V'(1)}, for all indices $i$ such that $x_i>1-\epsilon$, we can make the estimation that $V(x_i) \approx V(1)-(1-x_i)V'(1)= \frac{p-c}{1-\delta} - (1-x_i)\frac{p-q}{1-\delta}$. Let $i_{start}$ be the greatest index $i$ such that $x_{i_{start}}<1-\epsilon$. We have the following estimate: $i_{start} = O_x(a\cdot\log_{p/q}(1/\epsilon))$.

\begin{figure}[t]
 \centering
  \includegraphics[width=0.5\linewidth]{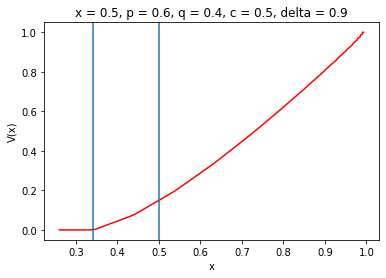}
  \caption{An example of computing $V(x)$, under dynamic prices, through our dynamic programming approach. Note that $x^* = 0.33$, which is lower than $x = 0.5 = \frac{c-q}{p-q}$, causes the local reward to be 0.}\label{plotDP}
  \end{figure}

Then, for all $i\le i_{start}$, we recursively compute $V(x_i)$, in decreasing order, with:
$$
V(x_i):= \frac{V(x_{i+b}) - R(x_{i+b}) - \delta\cdot\mathbb{P}_{x_{i+b}}(\like)V(x_{i+b+a})}{\delta \cdot\mathbb{P}_{x_{i+b}}(\dislike)}
$$
until we have $V(x_i)\le 0$. We call this index $i^*$ and then set $V(x_i):=0$ for all $i\le i^*$. The prior $x_{i^*}$ gives an estimation of $x^*$. This method is fast but it is difficult to prove an upper bound on the precision of the results obtained. \Cref{plotDP} presents an example that showcases the properties we have described. 

\paragraph{Static price scenario. } We can easily adapt this algorithm to compute an approximation of the global expected reward in the static price scenario. Here we have $R(x)=\pi-c$ which implies $V(1)=\frac{\pi-c}{1-\delta}$ and $V'(1)=0$. Then, we compute values $V(x_i)$ similarly, for all $i\le i_{start}$, until $x_i\le x_{\min}$. 

In the next section, we give a combinatorial method to compute $x^*$ and the global expected reward, that enables us to provide a strong guarantee on the solution.

\begin{figure}
\centering
    \includegraphics[width=0.6\linewidth]{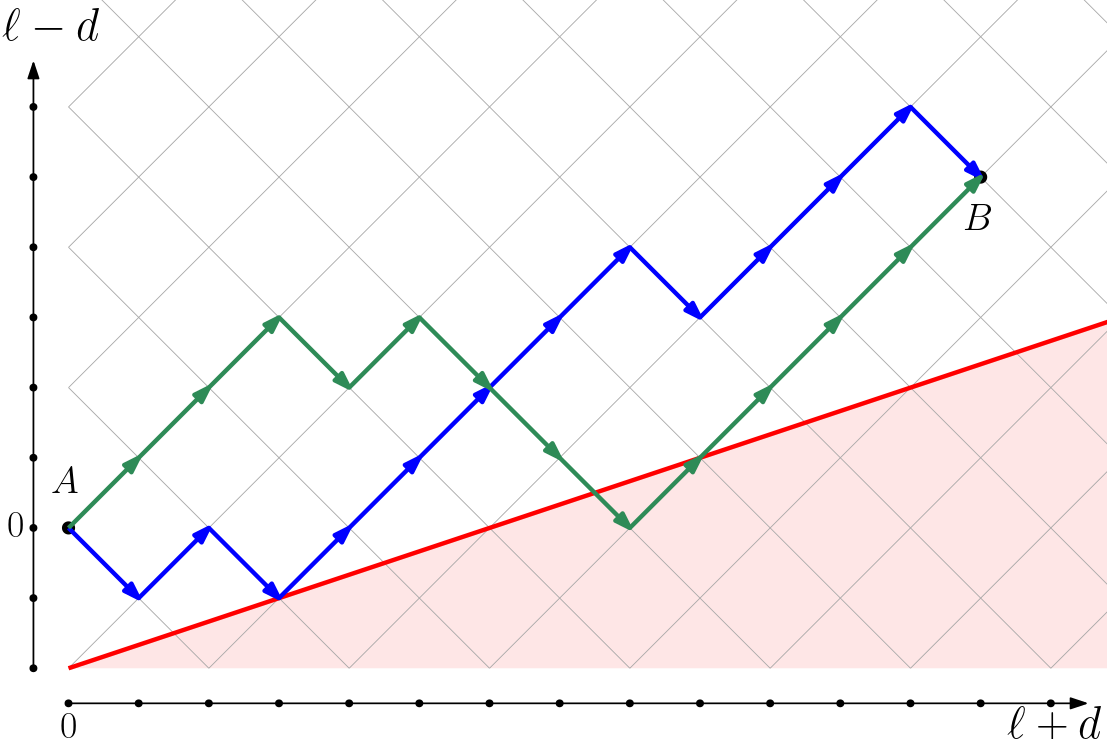}
    \caption{The number of paths from $A$ to $B$ along the grid, that do not enter the light red area (like the blue path but not like the green one) is the Catalan's quadrilateral number $C_3^{1,2}(9,4)=570$ (for comparison, the unconditional number of paths from $A$ to $B$ is ${{9+4} \choose 4}=715$ ). 
The slope of the boundary depends on parameters $a,b$. 
The classic Catalan's trapezoid 
arises when the boundary is horizontal.   }
    \label{fig:catalan}
\end{figure}

\section{A Combinatorial Approach}\label{section:combinatorial_approach}

In this section we give a good estimation of the value $V(x)$ in both pricing scenarios. 
To compute $V(x)$, we make use of the concepts of {\em Catalan's triangle} and {\em Catalan's trapezoid}. Conceptually, {\em Catalan's triangle} is a number triangle whose entries $C(\ell,d)$ correspond to the number of strings such that there are $\ell$ ``likes'' and $d$ ``dislikes'' such that no initial segment of the string has more dislikes than likes.  
The well-known \emph{Catalan numbers} correspond to $C(n,n)=\frac{1}{n+1} {2n \choose n}$. 

The so-called {\em Catalan's trapezoid} is an extension of Catalan's triangle, in which $C_m(\ell,d)$ counts the number of strings with $\ell$ likes and $d$ dislikes such that every initial segment has at least $m$ more likes than dislikes\footnote{We use here a slightly different definition than in \cite{R14}. We have $C_m'(\ell,d)=C_{m-1}(\ell,d)$ where $C_m'(\ell,d)$ denote the original Catalan Trapezoid numbers.}. In particular the Catalan's triangle corresponds to the special case where $m=0$.  
We have the following closed form for the Catalan's trapezoid \cite{R14}:
$C_m(\ell,d)={\ell+d \choose d}$ if $0\le d \le m$, $C_m(\ell,d)={\ell+d \choose d}-{\ell+d \choose d-m-1}$ if $m<d \le \ell + m$ and $C_m(\ell,d)=0$ otherwise. 

\begin{definition}[Catalan's quadrilateral]
Given integers $\ell,d$ and parameters $a,b$ and $m$ we denote $C_m^{a,b}(\ell,d)$ the number of strings consisting of $w$ L-s and $d$ D-s such that in every initial segment of the string that consists of $\ell'$ L-s and $d'$ D-s, the value $a\cdot \ell'-b\cdot d'$ is always at least $-m$. 
\end{definition}

Let us say that these numbers form \emph{Catalan's quadrilateral} since the Catalan's trapezoid corresponds to $a=b=1$. \Cref{fig:catalan} provides a geometrical interpretation of these numbers.

We have the following induction to compute these numbers: 
$C_m^{a,b}(\ell,d) = 0$ whenever $a\cdot \ell-b\cdot d < -m$. 
Otherwise $C_m^{a,b}(\ell,d) = 1$ when $\ell=0$ or $d=0$. And generally:
$
C_m^{a,b}(\ell,d) = C_m^{a,b}(\ell-1,d) + C_m^{a,b}(\ell,d-1)
$. 

Thus, computing $C_m^{a,b}(\ell,d)$ can be done in time $O(\ell\cdot d)$ with a simple dynamic program implementing the above inductive formulation. No non-recurrence-based formula exists for Catalan's quadrilaterals, though computation of partial values in the quadrilateral are derived in \cite{F13}.\footnote{See also \cite{HHP18} for some recent developments.}  
The purpose of defining these particular numbers lies in the following lemma that will enable us to provide an expression of the global expected reward. 

Given an initial prior $x$, we let $X_t$ be the (random) updated prior after $t$ reviews. In particular, let $X_0=x$. 
Recall that $x_{\ell,d}$ denotes the updated prior after a sequence of $\ell$ likes and $d$ dislikes from an initial prior $x$, and $R(x)$ is the local reward\footnote{Recall that $R(x)=\pi-c$ in the static price setting and $R(x)=xp+(1-x)q-c$ in the dynamic price setting} when the prior is $x$. The global expected prior from a prior $x$ is given by 
$$
V(x)=\sum_{{\ell,d\ge 0 , \\t=\ell+d}} \delta^{t}\cdot\mathbb{P}_x(X_{t} = x_{\ell,d}) \cdot R(x_{\ell,d}).
$$

We now use the Catalan's quadrilateral to provide an expression of $\mathbb{P}_x(X_{t} = x_{\ell,d})$. Let $x_\text{stop}$ denote the value of the prior when the selling process stops: in the dynamic pricing scenario, we have $x_\text{stop}=x^*$ and in the fixed price scenario, we have $x_\text{stop}=x_{min}=\frac{\pi-q}{p-q}$. 

\begin{lemma}
Let $\ell,d$ two integers, and $t=\ell+d$. Given any prior $x$, we have that
$$\mathbb{P}_x(X_{t} = x_{\ell,d}) = C_m^{a,b}(\ell,d)\cdot p_x(\ell,d),$$  
where 
$$p_x(\ell,d):= xp^\ell(1-p)^d + (1-x)q^\ell(1-q)^d
$$ 
is the probability of having a given ordered sequence of $\ell$ likes and $d$ dislikes; $a=\log(p/q)$; $b=\log(\frac{1-q}{1-p})$; and $m=\log_{\frac{1-q}{1-p}}\left(\frac{x(1-x_\text{stop})}{x_\text{stop}(1-x)}\right)$.
\label{lemma:proba_path}
\end{lemma}
\begin{proof}
    By \Cref{lemma:xwl}, the updated prior after a sequence of $\ell$ likes and $d$ dislikes only depends on $\ell$ and $d$ so as the probability of such each sequence. Therefore, $\mathbb{P}_x(X_{t} = x_{\ell,d})$ is the product of the probability of one sequence and the number of such sequences.

    We first show that the probability of having a given ordered sequence of $\ell$ likes and $d$ dislikes is 
    $$p_x(\ell,d)= xp^\ell(1-p)^d + (1-x)q^\ell(1-q)^d.
    $$ 
    We proceed by induction on the length $\ell+d$ of the sequence $\ell+d$. The base case of the induction follows from \Cref{lemma:xwl}. 
    Then, using the induction hypothesis, we obtain
    \begin{align*}
p_x(\ell+1,d) &= p_{L(x)}(\ell,d)\cdot\mathbb{P}_x(\like)
=(L(x)p^\ell(1-p)^d + (1-L(x))q^\ell(1-q)^d)
(xp+(1-x)q)\\
&=\Big(\frac{x p}{x p + (1-x) q}p^\ell(1-p)^d 
+(1-\frac{x p}{x p + (1-x) q})q^\ell(1-q)^d\Big)(xp+(1-x)q)\\ 
&= xp^{\ell+1}(1-p)^d + (1-x)q^{\ell+1}(1-q)^d. 
    \end{align*}
       
    The calculation for $p_x(\ell,d+1)=xp^{\ell}(1-p)^{d+1} + (1-x)q^{\ell}(1-q)^{d+1}$ works similarly. Thus we have established the induction. 
    
    Now for any integers $\ell,d$, it is easy to see that $x_{\ell,d}<x_\text{stop}$ if and only if $a\cdot \ell- b\cdot d < -m$ where $a=\log(p/q)$, $b=\log(\frac{1-q}{1-p})$ and $m=\log_{\frac{1-q}{1-p}}\left(\frac{x(1-x_\text{stop})}{x_\text{stop}(1-x)}\right)$. Thus, the number of sequences of $\ell$ likes and $d$ dislikes such that the prior at any time is at least $x_\text{stop}$ is equal to the Catalan's quadrilateral number $C_m^{a,b}(\ell,d)$. 
\end{proof}

In the dynamic pricing scenario, to compute a good approximation of $V(x)$, we first need to compute a good approximation of the threshold $x^*$. 

\paragraph{Computing $x^*$.}

By definition, $x^*$ is the prior for which stopping or continuing to play gives the same global expected reward. Assuming that the initial prior is $x=x^*$, we have $m=0$ and we obtain after simplification the following equation :
\begin{align*}
    0=V(x^*)
    =\sum_{\ell,d\ge 0} &\delta^{\ell+d} C_0^{a,b}(\ell,d)\cdot (x^*p^{\ell}(p-c)(1-p)^d
    +(1-x^*)q^{\ell}(q-c)(1-q)^d), 
\end{align*}
where $a=\log(p/q)$ and $b=\log(\frac{1-q}{1-p})$. 
If we set $\Phi(p):=\sum_{\ell,d\ge 0} \delta^{\ell+d} \cdot C_0^{a,b}(\ell,d)\cdot (p-c)p^{\ell}(1-p)^d$, the above equation becomes $0= x^*\Phi(p)+ (1-x^*)\Phi(q)$. Thus we can express $x^*$ as 
$$
x^*=\frac{\Phi(q)}{\Phi(q)-\Phi(p)}.
$$

To get a precise estimate of the value $\Phi(p)$, we only need to focus on sequences of likes and dislikes that do not exceed a certain length. More precisely, fix any $\epsilon>0$. Since $\Phi(p)$ is defined as a series of positive terms, we know that there exists an integer $t_\epsilon$ such that   
$$
\sum_{\ell,d\ge 0, \ell+d \ge t_\epsilon} \delta^{\ell+d} \cdot C_0^{a,b}(\ell,d)\cdot (p-c)p^{\ell}(1-p)^d\le \epsilon
$$
and since this series is upper bounded by a convergent geometric series, we have the following estimate $t_\epsilon=O(\log 1/\epsilon)$. Thus, we can compute an $\epsilon$-estimate $\widehat{x^*}$ of $x^*$, i.e. $|\widehat{x^*}-x^*|<\epsilon$, in time $O(\log(1/\epsilon)^2)$. When the ratio $a/b$ is a rational number, the set of possible updated priors from $x$ is discrete, so that choosing an $\epsilon$ sufficiently small enables to compute an exact value for $x^*$.  

\paragraph{Computing $V(x)$. } Once we have a precise estimation of $x^*$, we can proceed similarly to compute an arbitrarily close estimation of $V(x)$ for any prior $x$. For any $\epsilon>0$, there exists $t_\epsilon=O(\log 1/\epsilon)$, such that 
$$
\sum_{\ell,d\ge 0, \ell+d \ge t_\epsilon} \delta^{\ell+d} \cdot \mathbb{P}_x(X_{t} = x_{\ell,d}) \cdot R(x_{\ell,d})\le \epsilon.
$$ 
We can then use $x_\text{stop}$ and the values of the Catalan trapezoid to compute the sum $$
\hat{V}(x):=\sum_{{\ell,d\ge 0 , \\t=\ell+d\le t_\epsilon}} \delta^{t}\cdot\mathbb{P}_x(X_{t} = x_{\ell,d}) \cdot R(x_{\ell,d})
$$ and we have $|V(x)-\hat{V}(x)|\le \epsilon$.

In the \emph{symmetric} case, when the values of $p$ and $q$ are such that $q=1-p$, we can even get a closed expression for $x^*$ and $V(x)$. Indeed, we have $a=b=1$ and we can use the closed formula of the coefficients of the Catalan Trapezoid. 

\paragraph{Computing the optimal static price.} For each fixed price $\pi$, we can thus compute the global expected reward for the seller. In order to maximize her revenue, the seller can optimize this function over the values of $\pi\in (c,xp+(1-x)q]$. Clearly $\pi$ has to be at least $c$ for the revenue to be positive. Also if $\pi>xp+(1-x)q$ then no user will ever buy and then the revenue is zero.

\begin{figure}
  \centering
  \includegraphics[width=.48\linewidth]{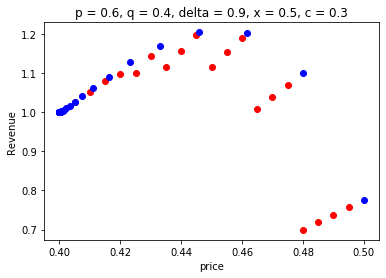} 
  \includegraphics[width=.48\linewidth]{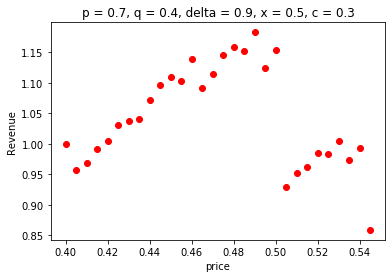}
  \caption{The value of global expected reward depending on the fixed price $\pi$. In the symmetric setting (left), the blue points represent the revenue on the efficient frontier, which is the maximum possible price per discrete $x_\text{stop}$. The red points are not optimal because such prices yield the same number of possible net dislikes before the buyers stop buying. In the general setting (right), we see that revenue as a function of price is not as well-defined.}
  \label{fig:static}
\end{figure}

In the symmetric setting, note that for a  price $\pi$, buyers can initially tolerate up to a net of $m_\pi$ dislikes, where $m_\pi$ is a function decreasing in $\pi$ before no longer buying. Thus, for a fixed integer $m$, we can maximize the global expected reward by setting the maximum price $\pi$ such that $m_\pi=m$. This efficient frontier appears to be concave, as shown in  \Cref{fig:static}, so finding the optimal price is simply a binary search procedure along the efficient frontier. In the asymmetric setting, however, $m_\pi$ is a function of both the number of likes and dislikes, due to the behavior of Catalan's quadrilateral. \Cref{fig:static}, shows that computing the optimal price in the asymmetric case requires searching over a larger set of prices. 

\section{Success and Failure of Learning}
\label{sec:comparison}

In this section we investigate the probability that the market learns the true value of the product, i.e. the probability of stopping in finite time when the product is bad and the probability of selling forever when the product is good. If dynamic pricing is used, learning occurs with larger probability. Our main conclusion is to express this additional gain as a function of the primitives of the model, proving a simple quantification of the potential gains of dynamic pricing over static pricing. 

A first observation 
is that in both models the market will always discover if the product is bad. Recall that we assume that $q\le c$. 

\begin{lemma}
Assuming that the product is bad, we stop selling in finite time almost surely.
\label{lemma:stopwhenbad}
\end{lemma}
\begin{proof}
    Let $D_t=a\ell_t-bd_t$ denote the random variable that corresponds to the weighted difference between the number of likes and dislikes after $t$ reviews, where $a=\log (p/q)$ and $b=\log((1-q)/(1-p))$. We define the stopping time $\tau$ as the first time $t$ when $D_t<-m$ where $m$ depends on the original prior $x$ and the threshold prior $x_\text{stop}$. 
    
    %and assume that $d_0=0$.
    Given that the product is bad, we have:
    $\mathbb{E}(D_t - D_{t-1})=a\cdot\mathbb{P}(\like|\text{bad})-b\cdot\mathbb{P}(\dislike |\text{bad}) =aq-b(1-q)=:\mu <0$ for any $0<q<p$. Then, $\mathbb{E}(D_t)=\mu\cdot t\rightarrow_t -\infty$. 
    
    We deduce, for $t$ sufficiently large, using Bienaymé-Tchebychev inequality, that 
    $$
    \mathbb{P}(\tau \ge t)\le \mathbb{P}\left(|D_t-\mathbb{E}(D_t)|\ge -\mu\cdot t-m\right)\le O(1/t)
   $$
where $x_\text{stop}=x^*$ for the dynamic pricing model and $x_\text{stop}=x_{\min} $ for the single price model.\footnote{We now give an exact expression of this probability in the symmetric case, in  \Cref{lemma:exact_expr}.}
\end{proof}

On the other hand, when the product is good, learning may fail to occur. To quantify this efficiency loss, recall that $x_\text{stop}$ corresponds to the threshold prior from which the users stop buying. In the dynamic pricing model we have $x_\text{stop}=x^*$ and in the static price model, we have $x_\text{stop}=x_{\min}=\frac{\pi-q}{p-q}$. 
The main result of this section,  \Cref{lemma:estimation_playing_forever}, establishes that when the product is good the probability of learning it is at least $\frac{x-x_\text{stop}}{x(1-x_\text{stop})}$. 

Given an initial prior $x$, for all time $t\ge 0$ we define the (random) variable $X_t$ that is the updated prior after $t$ reviews.  $(X_t)_{t\ge 0}$ is a martingale with $X_0=x$. Now, let $\tau$ denote the (random) time at which the selling process stops. 

\begin{lemma}
Given an initial prior $x$, we have the following estimation on the probability that the process continues forever: $\frac{x-x_\text{stop}}{1-x_\text{stop}}< \mathbb{P}_{x}(X_\tau=\infty)\le\frac{x-D(x_\text{stop})}{1-D(x_\text{stop})}$. Additionally, if we assume that the product is good, then the probability of learning that the product is good is $\mathbb{P}_{x}(X_\tau=\infty \mid \text{good})> \frac{x-x_\text{stop}}{x(1-x_\text{stop})}$. 
\label{lemma:estimation_playing_forever}
\end{lemma}
\begin{proof}
    Let us fix $\epsilon>0$. The random time $\tau_\epsilon$ at which $X_t$ reaches $x_\text{stop}$ or $1-\epsilon$ is a stopping time. Since $\tau_\epsilon$ has finite expectation, by the optional stopping theorem, the expected value of $X_{\tau_\epsilon}$ is equal to the initial prior, i.e., $\mathbb{E}(X_{\tau_\epsilon})=x$. Then we get 
    \begin{align*}
    x=\mathbb{P}_{x}&(X_{\tau_\epsilon}<x_\text{stop})\mathbb{E}(X_{\tau_\epsilon} | X_{\tau_\epsilon} <x_\text{stop})
    + \mathbb{P}_{x}(X_{\tau_\epsilon}>1-\epsilon)\mathbb{E}(X_{\tau_\epsilon} | X_{\tau_\epsilon}>1-\epsilon).
    \end{align*}
    We know that $D(x_\text{stop})\le\mathbb{E}(X_{\tau_\epsilon} | X_{\tau_\epsilon} <x_\text{stop})< x_\text{stop}$ and $1-\epsilon< \mathbb{E}(X_{\tau_\epsilon} | X_{\tau_\epsilon}>1-\epsilon) \le L(1-\epsilon)$.

Thus, when $\epsilon$ goes to zero, we obtain $\mathbb{P}_{x}({\tau}=\infty)=\frac{x-\mathbb{E}(X_\tau | X_\tau<x_\text{stop})}{1-\mathbb{E}(X_\tau | X_\tau<x_\text{stop})}$.  The estimation then follows from the fact that $D(x_\text{stop})\le\mathbb{E}(X_\tau \mid X_\tau<x_\text{stop})< x_\text{stop}$. 

To prove the second part of the statement we simply use: 
$$
\mathbb{P}_x(\tau = \infty)=\mathbb{P}_x(\tau = \infty \mid \text{good})\cdot \mathbb{P}_x(\text{good})+ \mathbb{P}_x(\tau = \infty \mid \text{bad})\cdot \mathbb{P}_x(\text{bad})=\mathbb{P}_x(\tau = \infty \mid \text{good})\cdot x
$$ 
    where $\mathbb{P}_x(\tau = \infty \mid \text{bad})=0$ holds by \Cref{lemma:stopwhenbad}. 
\end{proof}

Notice that since $x^*<x_{\min}$, we will learn that the product is bad in the static pricing scenario earlier than in the dynamic pricing model. Conversely, we learn that the product is good with higher probability in the dynamic pricing model. 

In the symmetric case, i.e. when $q=1-p$, we can have a better estimation. Indeed, if $\tau=t$ then necessarily, $X_{t-1}=x_\text{stop}$ and we observe a dislike at time $t-1$. Thus, $\mathbb{E}(X_\tau | X_\tau <\text{stop})=D(x_\text{stop})$ so that $\mathbb{P}_{x}(X_\tau=\infty)=\frac{x-D(x_\text{stop})}{1-D(x_\text{stop})}$. See also \Cref{lemma:exact_expr} in the Appendix.

With these results we can bound the ratio of not learning under the considered pricing strategies. This happens exactly when the product is good but the market does not discover it and stops buying in finite time. Let $FN_\text{static}$ and $FN_\text{dynamic}$ be the probabilities of stopping when the product is good in the static and in the dynamic prices scenarios, respectively. In the case when $p=1-q$ we have $\frac{FN_\text{static}}{FN_\text{dynamic}}=\frac{1/D(x^*)-1}{1/D(x_{\min})-1}$; see \Cref{lemma:FN} in the Appendix. 

\section{Extension}\label{sec:extension}

We now consider the problem when the product has a quality $q\in Q\subseteq [0,1]$. Again, a product has quality $q\in Q$ if it is liked by a $q$ fraction of people, or equivalently, if the probability that a given person likes the product is $q$. In this model, the prior on the quality of the product becomes a random variable $X\in Q$. 
Given the current prior $X$, we update the prior to $L(X)$ after a like ($\like$) and to $D(X)$ after a dislike ($\dislike$) as follows:
\begin{align*}
&\P(L(X)=q):=\P_X(X=q\mid \like) =\frac{\P_X(\like\mid X=q)\P_X(X=q)}{\P_X(\like)} = \frac{q}{\E(X)}\P(X=q);
\end{align*}
\begin{align*}
&\P(D(X)=q):=\P_X(X=q\mid \dislike)=\frac{\P_X(\dislike\mid X=q)\P_X(X=q)}{\P_X(\dislike)} = \frac{1-q}{1-\E(X)}\P(X=q).
\end{align*}
These equations correspond to the case in which $Q$ is a discrete set and should be replaced by the corresponding probability density functions in case $Q$ is a continuous set. 
Again, the prior after a sequence of likes and dislikes is independent, of the order in which the likes and dislikes happened. Therefore we have a simple expression of the prior after a given sequences of reviews. 

\begin{lemma} \label{lemma:xwl_general}
    Given a prior $X$, the distribution of the updated prior $X_{\ell,d}$ after a sequence of $\ell$ likes and $d$ dislikes is given by: 
$$\P(X_{\ell,d}\le q)= \frac{\E(X^\ell(1-X)^d|X\le q) \P(X\le q)}{\E(X^\ell(1-X)^d)}, $$
for all  $q\in Q$.
\end{lemma}
\begin{proof}
By iterating the calculations above for the distribution of $L(X)$ and $D(X)$ one can obtain that $\P(X_{\ell,d}=q)=\frac{q^\ell(1-q)^d}{\E(X^\ell(1-X)^d)}\P(X=q)$, for any $q\in Q$. Of course, if $Q$ is a continuous set, we need to replace the probabilities by the corresponding densities. 
To conclude the lemma, we simply need to integrate this equation for possible quality values in $Q$ that are below $q$.
\end{proof}

Initially, suppose that the quality prior is $X_0$. At any time, both the seller and the buyers have access to the current prior $X$. So the buyer will buy the product if its utility $\E(X)-\pi$ is non-negative, where $\pi$ is the current price of the product. In the dynamic price setting, in order to maximize her revenue, the seller must set the product at price $\pi:= \E(X)$. Of course, at any point in time, the seller may decide to stop selling the product, and she will do this if the future expected reward is negative. With this, we can again write a dynamic program to estimate the seller's optimal revenue under a dynamic pricing strategy. First note that given a prior $X$, the value of the total expected reward (for the seller) satisfies the following equation: 
\begin{align}
     V(X)=\max&\Big( 0,\E(X)-c+ \delta\cdot\big(\P_X(\like)V(L(X))
     + \P_X(\dislike)V(D(X))\big)\Big),
    \label{eq:recurrence_general}
\end{align}
where $c$ is the production cost per unit.  
Define the estimator $\widehat{V}(X):=\frac{\E(X)-c}{1-\delta}$ that does not take into account the reviews. We now describe a dynamic programming approach to compute an approximation $\widetilde{V}(X)$ of $V(X)$. For this, we consider a constant $M$, to be specified later. We set $\widetilde{V}(X_{\ell,d})=\widehat{V}(X_{\ell,d})$, for all $\ell,d$ such that $\ell+d=M$. (As we note later, for $M$ relatively large and a well behaved initial prior distribution $X_0$, the random variable $X_{\ell,d}$ is highly concentrated around its mean, which is roughly $\ell/(\ell+d)$, and thus $\widehat{V}(X_{\ell,d})$ is very close to $V(X_{\ell,d})$). 
For $i=M-1,\dots, 0$ and all $\ell,d$ such that $\ell+d=i$, we compute $\widetilde{V}(X_{\ell,d})$ using \eqref{eq:recurrence_general}, where we have replaced $V$ by $\widetilde{V}$. This can be done in time $O(M^2)$. 

\begin{proposition}
$|\widetilde{V}(X)-V(X)|\le \frac{1-c}{1-\delta}\delta^M$. 
\label{claim:good_estimate}
\end{proposition}
\begin{proof}
    We prove by induction that $|\widetilde{V}(X_{\ell,d})-V(X_{\ell,d})|\le \frac{1-c}{1-\delta}\delta^{M-i}$, where $i=\ell+d$. 
    We trivially have $|\widetilde{V}(X_{\ell,d})-V(X_{\ell,d})|\le \frac{1-c}{1-\delta}$ when $\ell+d=M$. Now, when $i=\ell+d<M$, we have
    \begin{align}
    &|\widetilde{V}(X_{\ell,d})-V(X_{\ell,d})|\nonumber\\
    &\le\delta\cdot\Big(\P_{X_{\ell,d}}(\like)|\widetilde{V}(X_{\ell+1,d})-V(X_{\ell+1,d})|
    +\P_{X_{\ell,d}}(\dislike)|\widetilde{V}(X_{\ell,d+1})-V(X_{\ell,d+1})|\Big)\nonumber\\
    &\le \delta\Big(\P_{X_{\ell,d}}(\like)\frac{1-c}{1-\delta}\delta^{M-(i+1)}
    +\P_{X_{\ell,d}}(\dislike)\frac{1-c}{1-\delta}\delta^{M-(i+1)}\Big)=
    \frac{1-c}{1-\delta}\delta^{M-i}\nonumber.
    \end{align}   
\end{proof}

\begin{figure}[t]
    \centering
    \includegraphics[width=.4\linewidth]{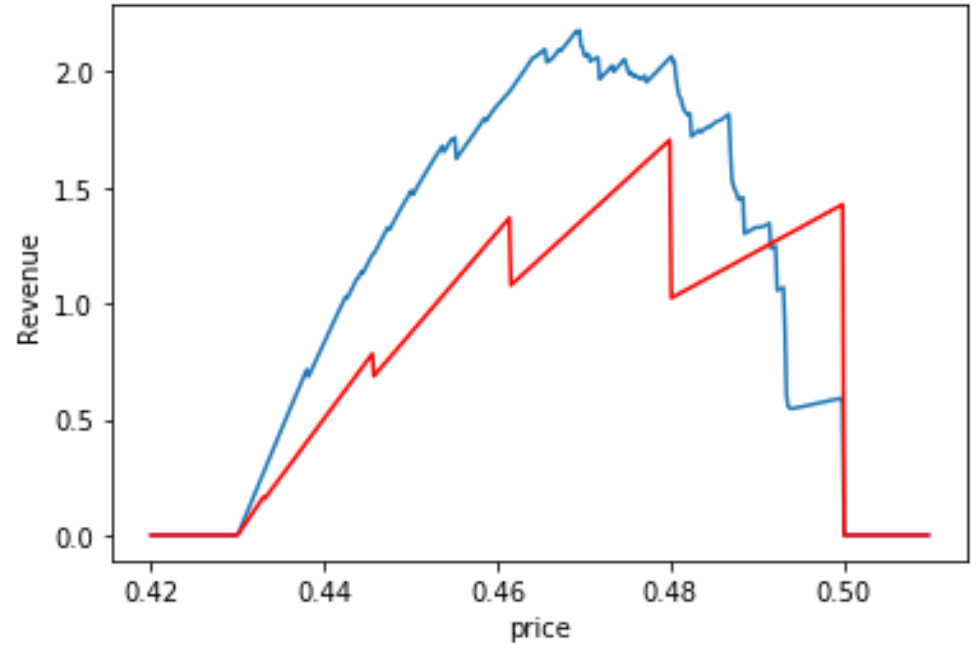}
    \caption{Seller's revenue depending on the static price. The red plot corresponds to the good/bad (binary) model with $p = 0.6, q = 0.4, c = 0.43, \delta = 0.99$, and the prior is such that the product is good/bad with probability 0.5. The blue plot corresponds to the extended model with $Q=[0.4, 0.6]$ and  uniform prior distribution.\vspace{-2ex}}
    \label{fig:fixed_price}
\end{figure}

The previous result provides a general bound on the quality of our dynamic programming approach. When the initial prior is discrete, the problem is combinatorial and one can use similar ideas from \Cref{section:combinatorial_approach} to obtain improved guarantees. On the other hand, when the set of possible values is continuous, we can obtain significantly better bounds with additional properties on the initial prior. In particular, when the initial prior admits a continuous and strictly increasing distribution with bounded density, we have that for $\ell+d$ large, $X_{\ell,d}$ is highly concentrated around $\ell/(\ell+d)$. For instance, its variance is $Var(X_{\ell,d})=O(\frac{1}{\ell+d})$. 
Additionally, if $\E(X)$ is bounded away from $c$, using Bienaymé-Tchebychev inequality (or simply the law of large numbers), one can show that the probability that the updated prior $X'$ obtained after $M$ reviews is such that $\E(X')<c$ is $O(1/M^2)$. With this, the bound of  \Cref{claim:good_estimate} improves to $O(\frac{\delta^M}{M(1-\delta)})$. 

In terms of static pricing, we can easily adapt the approach to this more general model. Again, the seller wants to fix a price $\pi$, and the buyer buys if $\E(X)\ge \pi$, where $X$ is the public prior distribution. Based on this, we can devise a simple dynamic program similar to the algorithm of  \Cref{claim:good_estimate}, to obtain a good estimation of the total revenue for the seller. Again, we can prove that if the product is bad, i.e., the true value of the product is smaller than the cost, then the process will stop almost surely. In \Cref{fig:fixed_price} we plot the revenue as a function of the static price in both models, setting parameters to make them comparable (equal expected quality and equal range of product quality).

\begin{figure}
    \centering
    \includegraphics[width=.45\linewidth]{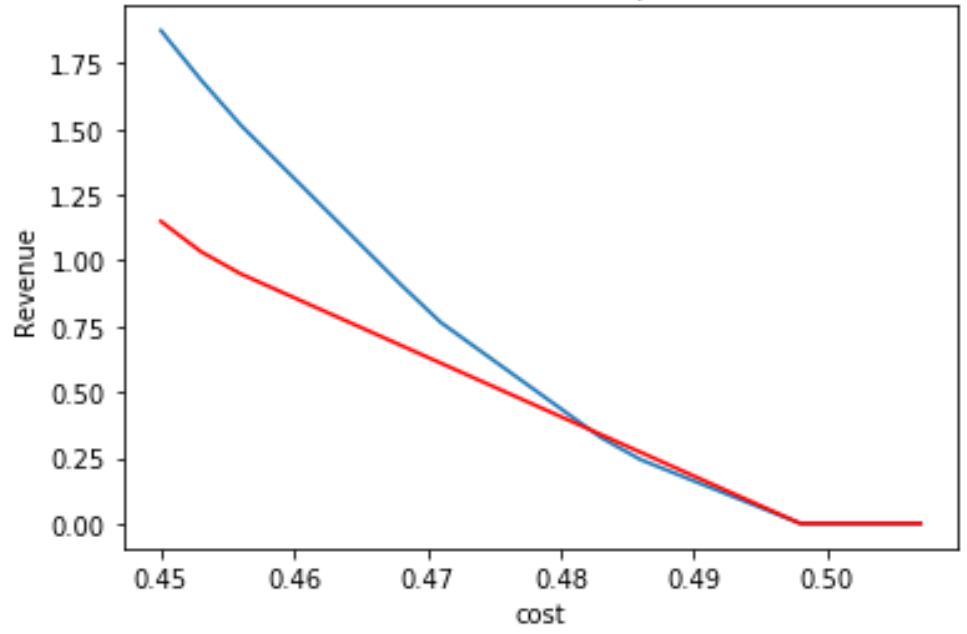}
    \includegraphics[width=.44\linewidth]{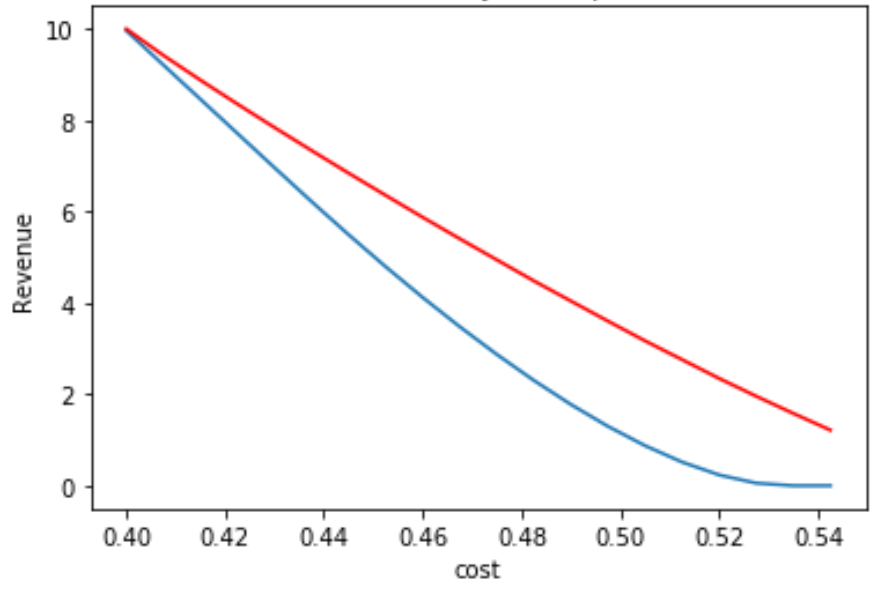}
    \caption{Seller's revenue as a function of the cost in the same instances as in \Cref{fig:fixed_price}. The red function corresponds to the good/bad model, and the blue function corresponds to the extended model. On the left, we plot revenues resulting from the optimal static pricing (the price is optimized for each possible cost), and on the right, we plot revenues resulting from dynamic pricing.\vspace{-2ex}}
    \label{fig:fixed_cost}
\end{figure}

Finally, we evaluate the expected revenue of both the static and dynamic pricing policies in both models in the comparable setting just described. The situation is depicted in  \Cref{fig:fixed_cost}. A surprising result is that while for the optimal static price, the seller's revenue in the extended model is mostly higher than that in the binary model, the situation changes dramatically for dynamic pricing. The intuition is that in dynamic pricing, if $c>0.4$, then the expected quality minus $c$ is higher in the good/bad model. However, when considering static prices this effect is less relevant than the faster learning process that occurs in the extended model. Indeed, in that case, if $c$ is relatively small, then the extended model gets higher revenue since the product is more interesting, and the buyers quickly learn that it is worth buying. However, as the cost approaches 0.5, the product becomes less interesting and then the faster learning implies that the seller's revenue decreases faster in the extended model.

\bibliographystyle{plain}
\bibliography{bibliography}

\appendix
\section*{Appendix}

\begin{lemma}
Assume that $q=1-p$. Then, when the product is good, the probability of selling forever is equal to $1-\left(\frac{1-\sqrt{1-4p(1-p)}}{2p}\right)^{m}$ where $m$ is the smallest integer such that after $m$ dislikes the prior goes below $x_\text{stop}$. 
\label{lemma:exact_expr}
\end{lemma}

\begin{proof}(\Cref{lemma:exact_expr})
    Let $p_m$ denote this probability. We have $p_m=0$ if $m\le 0$; $p_m=p\cdot p_{m+1}+(1-p)\cdot p_{m-1}$ otherwise, and $\lim_m p_m = 1$. The roots of the polynomial $pX^2-X+(1-p)$ are $1$ and $\frac{1-\sqrt{1-4p(1-p)}}{2p}<1$. Thus, we deduce easily the expected expression. 
\end{proof}

\begin{lemma}
Let $FN_\text{static}$ and $FN_\text{dynamic}$ be the probabilities of stopping when the product is good, respectively in the static and in the dynamic prices scenarios.   We have $FN_\text{static}\ge FN_\text{dynamic} > 0$ and in the case when $p=1-q$:
$$
\frac{FN_\text{static}}{FN_\text{dynamic}}=\frac{D(x_{\min})}{D(x^*)}\cdot\frac{1-D(x^*)}{1-D(x_{\min})}=\frac{1/D(x^*)-1}{1/D(x_{\min})-1}.
$$
\label{lemma:FN}
\end{lemma}
\begin{proof}(\Cref{lemma:FN})
    By \Cref{lemma:estimation_playing_forever}, in the symmetric case, the probability of having a false negative is $\frac{D(x_\text{stop})}{1-D(x_\text{stop})}\cdot\frac{1-x}{x}$ and the formula follows.
\end{proof}

\end{document}